\pdfoutput=1

\documentclass[11pt]{article}

\usepackage[final]{acl}

\usepackage{times}
\usepackage{latexsym}

\usepackage[T1]{fontenc}

\usepackage[utf8]{inputenc}

\usepackage{microtype}

\usepackage{inconsolata}

\usepackage{graphicx}
\usepackage[capitalize,noabbrev]{cleveref}
\usepackage{xspace}
\newcommand{\teamname}{\textsc{KnowComp Pokemon}\xspace}

\title{\teamname Team at DialAM-2024: A Two-Stage Pipeline for Detecting Relations in Dialogical Argument Mining}

\author{
 \textbf{Zihao Zheng\textsuperscript{1}},
 \textbf{Zhaowei Wang\textsuperscript{2}},
 \textbf{Qing Zong\textsuperscript{2}},
 \textbf{Yangqiu Song\textsuperscript{2}},
\\
 \textsuperscript{1}Harbin Institute of Technology(Shenzhen), Guangdong, China \\
 \textsuperscript{2}Department of Computer Science and Engineering, HKUST, Hong Kong SAR, China
\\
\texttt{
   \{melfeszheng, zongqing0068\}@gmail.com, 
   \{zwanggy, yqsong\}@cse.ust.hk
 }
}

\begin{document}
\maketitle
\begin{abstract}
Dialogical Argument Mining (\textbf{DialAM}) is an important branch of Argument Mining (\textbf{AM}). DialAM-2024 is a shared task focusing on dialogical argument mining, which requires us to identify argumentative relations and illocutionary relations among proposition nodes and locution nodes. To accomplish this, we propose a two-stage pipeline\footnote{Codes are avilable at https://github.com/HKUST-KnowComp/KnowComp-DialAM2024-ACL2024}, which includes the Two-Step S-Node Prediction Model in Stage 1 and the YA-Node Prediction Model in Stage 2. We also augment the training data in both stages and introduce context in Stage 2. We successfully completed the task and achieved good results. Our team \textbf{\teamname} ranked \textbf{1\emph{st}} in the ARI Focused score and \textbf{4\emph{th}} in the Global Focused score.

\end{abstract}

\section{Introduction}

Dialogues contain a wealth of information about arguments and their relationships, but the structure and content of dialogues are casual, which poses challenges for extracting argument structures. To handle it, \citet{IAT-dialogue} provides a method for analyzing dialogue and argument structures, as well as the relations between them, using Inference Anchoring Theory (IAT)~\citep{IAT}. In dialogues, the content of the discussions serves as locution nodes, while their propositional content serves as proposition nodes. Among these nodes, three types of relation nodes are used for connection: argumentative relations between propositions, illocutionary relations between locutions and propositions, and transitional relations between locutions. This method helps extract argument structures from dialogues, enabling further argument mining and analysis. By employing this approach, \citet{qt30} has introduced QT30, an English corpus of meticulously analyzed dialogical argumentation. This corpus encompasses the argumentative structure derived from 30 debates from the BBC television program Question Time.

The DialAM task in ACL2024~\citep{ruizdolz-2024-overview} is the first shared task focused on dialogical argument mining. It consists of two tasks. The first task is to identify Propositional Relations, aiming to detect argumentative relations between the identified and segmented propositions in the argumentative dialogue. The second task is the Identification of Illocutionary Relations, which aims to detect the illocutionary relations between the locutions uttered in the dialogue and the argumentative propositions associated with them. 

To address the two tasks proposed by DialAM-2024, we introduce a two-stage pipeline. Based on initial locutions and propositional contents, we utilize data augmentation by adding data that does not fit any relation in the relation set to increase the gap between data within and outside the relation set. 
Thus, we can predict the relationships between propositional contents using our proposed two-step S-node prediction model to address the first task. Building upon this, we further tackle the task of identifying illocutionary relations by bringing context to prediction and employing a multi-classification YA-node prediction model. Adopting this method, our team \textbf{Pokemon} ranked \textbf{1\emph{st}} in the ARI Focused score and \textbf{4\emph{th}} in the Global Focused score.

Our paper is structured as follows: Section 2 presents related work on argument mining. Section 3 describes the details of our proposed method, a two-stage pipeline. Section 4 outlines the experiments we conducted, including the models and methods used in each stage, as well as the overall pipeline experiments. Section 5 makes a conclusion and provides further discussion.

\begin{figure*}[ht]
  \centering
  \includegraphics [width=\textwidth]{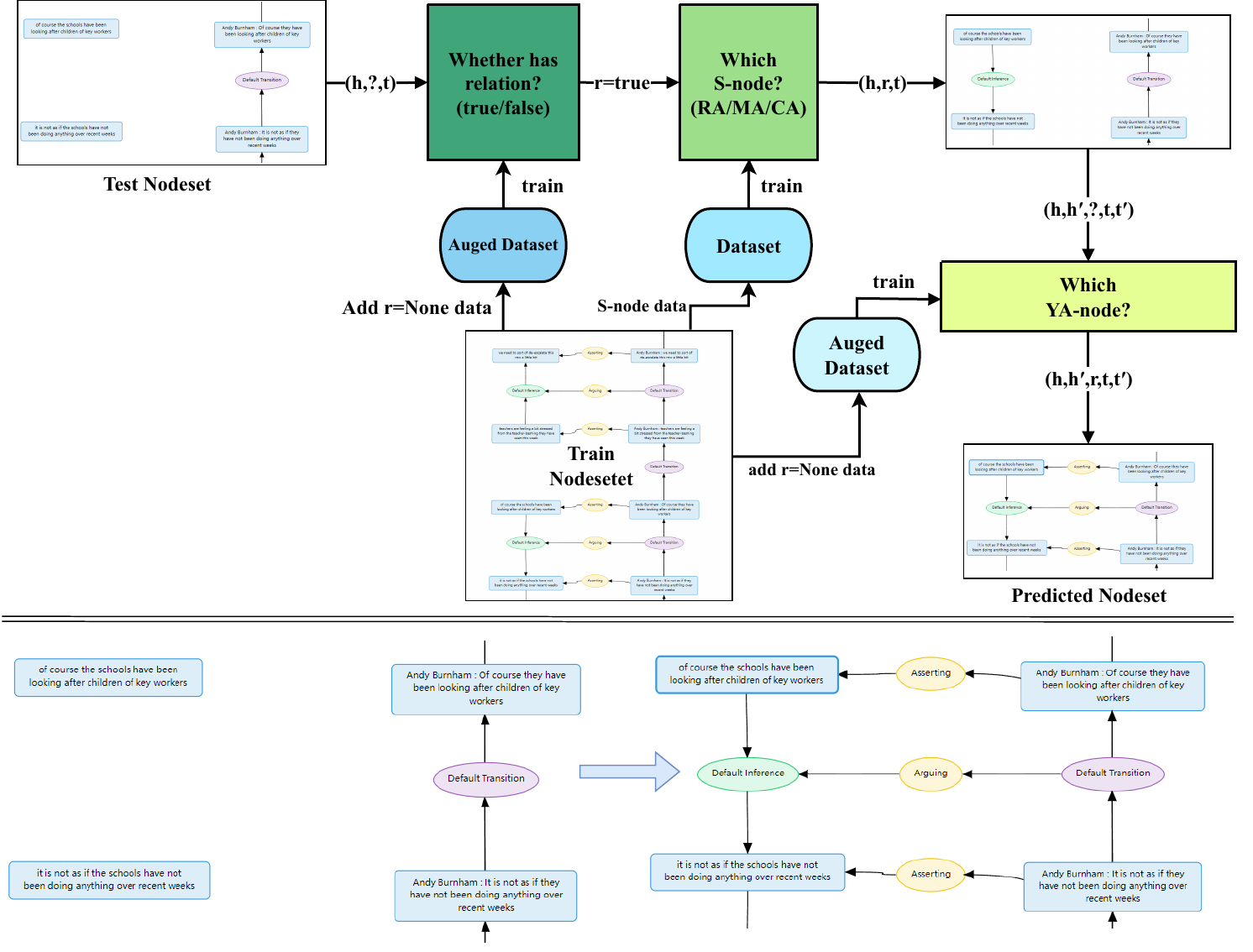}
  \caption{The 2-stage Pipeline.}
  \label{fig:pipeline}
\end{figure*}

\section{Related Work}
\paragraph{Argument Mining:} Argument Mining involves the automatic extraction and analysis of arguments from various sources, such as texts, debates, and social media discussions~\citep{stab-gurevych-2014-annotating, habernal-gurevych-2017-argumentation, carlile-etal-2018-give, lawrence-reed-2019-argument}. Some recent works study the stance and persuasiveness of the arguments in multi-modal data like tweets on Twitter~\citep{liu-etal-2022-imagearg, tilfa}. Other works focus on dialogical argumentation, exploring how arguments are put forward, supported, and attacked through dialogue~\citep{Haddadan, Visser}. QT30 corpus~\citep{qt30}, which is built on Inference Anchoring Theory (IAT)~\citep{IAT}, a prominent framework in manual argument analysis, is the largest dialogical argumentation corpus in English.
\newpage

\section{Method}
We have developed a pipeline (Fig.~\ref{fig:pipeline}) to address the challenge of dialogical argument mining. This pipeline consists of two stages designed to address the task of identifying propositional relations and illocutionary relations, respectively.

\subsection{Two-Step S-node Prediction Model}
\label{section3_1}
Our primary objective in the first stage is to detect argumentative relations between propositions (I-node). According to QT30~\citep{qt30}, This kind of relation (S-node) consists of Inference (RA-node), Rephrase (MA-node), and Conflict (CA-node). However, it is worth noting that not all I-node pairs have relations. Consequently, an initial determination should be made regarding the presence of a relation between two given I-nodes, followed by a secondary prediction of the specific scheme of the relation. This binary step-wise approach forms the foundation of our two-step prediction model.

Inspired by the approach proposed by \citet{parikh-etal-2016-decomposable}, we adopt a similar representation using pairs to denote our problems. Specifically, for any two distinct I-nodes denoted as $h$ and $t$, wherein $h$ represents the head node and $t$ the tail node, the task is to predict the relation $r$ between $h$ and $t$ given the tuple $(h, t)$ and subsequently deriving the final triple $(h, r, t)$.

The first step of determining relation existence is framed as a binary classification task, given the pair $(h, t)$, with the relation set $R = \{true, false\}$. The principle of cross-entropy loss shapes the loss function of the model. 


Similarly, the second step of ascertaining the specific relation between the I-nodes is structured as a ternary classification task, with the relation set $R = \{RA, CA, MA\}$. 

\subsection{YA-node Prediction Model}
\label{section3_2}
The illocutionary relations (YA-node) include (11 distinct types in total): 1) Asserting, Challenging, Pure Questioning, Assertive Questioning, Rhetorical Questioning between I-nodes and L-nodes, 2) Arguing, Disagreeing, Default Illocuting, Restating between TA-nodes and S-nodes, and 3) Agreeing, Challenging, Disagreeing between TA-nodes and I-nodes~\citep{qt30}. 
The relationship between L-node and I-node is relatively direct, indicating an illocutionary relation between locutions and their propositional content. However, for the occasion where YA-nodes are connected to TA-nodes or S-nodes, since TA-nodes and S-nodes themselves do not have much meaning when considered alone, we take the context into account, that is, considering two L-nodes connected by TA-nodes and two or more I-nodes connected by S-nodes.

Our task still remains to predict the relation $r$ between the given head node $h$ and tail node $t$. Additionally, the head and tail nodes may be followed by their respective contexts $h’$ and $t’$.

This is also a multi-classification task to predict the illocutionary relation $r$ given $(h, h’, t, t’)$.  The relation set $R=\{r_0,r_1,r_2,...,r_{11}\}$, where $r_0$ indicates there’s no illocutionary relation between the node pairs. The model's loss function is cross-entropy loss.

\subsection{Data Augmentation}
While we have discussed the pipeline of our framework in the above two sections (i.e., \cref{section3_1} and \cref{section3_2}), we also introduced data augmentation techniques to further improve the performance of fine-tuned models in our framework. 

Within the training dataset of the first step of the first stage, I-node pairs already connected by S-nodes are categorized as $r=true$. It becomes imperative to introduce $r=false$ data manually. To this end, a set number of I-node pairs without S-node connections are randomly selected to represent the training data for $r=false$. 
Specifically, in each nodeset within our training set, we randomly select some node pairs from all possible I-node pairs. These selected I-node pairs must satisfy the condition that there is no S-node connecting them. We think that there are no significant argumentative relations between these selected I-node pairs. 
Meanwhile, the training dataset for the second step is solely comprised of I-node pairs with established S-node connections, but the connections are further categorized into $RA$, $MA$, and $CA$.

In the training set of the YA-node prediction model of the second stage, in addition to the tuples $(h, h’, r_{1-11}, t, t’)$ that already have YA-node connections as training data, a certain number of tuples $(h, h’, r_0, t, t’)$ need to be extracted from node pairs that do not have YA node connections, artificially created as training data with $r=r_0$, i.e., $r=None$.

\section{Experiments}
\subsection{Setup}
The baseline models we employed include DeBERTa-base~\citep{he2021debertav3}, DeBERTa-large,  DeBERTa-MNLI, RoBERTa-MNLI~\citep{liu2019roberta}. We also tried LLaMa-3-8B~\citep{llama3modelcard} with LoRa~\citep{hu2022lora}.

The learning rate during training is 1e-5, the weight decay is 0.01, and fp16 is enabled during the training process. When utilizing Lora, the parameter r is set to 64, and alpha is set to 16. Due to time constraints, the testing of other LoRa parameters was not completed.

Our dataset comprises a total of 1,478 nodesets. We randomly selected 78 nodesets as the evaluation set, leaving the remaining 1,400 nodesets for the training set. A more detailed data description is in appendix~\ref{sec:data_desc}.

\begin{table*}[ht]
  \centering
  \begin{tabular}{c|ccc|ccc}
    \hline
    \textbf{Model} & \multicolumn{3}{c|}{\textbf{General Metrics}} & \multicolumn{3}{c}{\textbf{Focused Metrics}} \\
    & \textbf{precision} & \textbf{recall} & \textbf{f1} & \textbf{precision} & \textbf{recall} & \textbf{f1} \\
    \hline
    RoBERTa-MNLI         & 0.114 & 0.369 & 0.046 & 0.494 & 0.533 & 0.488 \\
    DeBERTa-large     & 0.099 & 0.376 & 0.050 & 0.511 & 0.548 & 0.503 \\
    LLaMa-3-8B-LoRa       & 0.100 & 0.289 & 0.018 & 0.261 & 0.432 & 0.315 \\
    \hline
    DeBERTa-large       & 0.351 & 0.443 & 0.322 & 0.351 & 0.266 & 0.282 \\
    RoBERTa-MNLI       & 0.317 & 0.470 & 0.306 & 0.449 & 0.334 & 0.355 \\
    \hline
  \end{tabular}
  \caption{Experiments on different methods of the first stage of S-node prediction. The two models in the lower part of the table are the 2nd-step models, while the four models in the upper part are four-label classification models. }
  \label{tab:comparison_on_stage1}
\end{table*}

\begin{table*}[ht]
  \centering
  \begin{tabular}{c|ccc|ccc}
    \hline
    \textbf{Model} & \multicolumn{3}{c|}{\textbf{General Metrics}} & \multicolumn{3}{c}{\textbf{Focused Metrics}} \\
    & \textbf{precision} & \textbf{recall} & \textbf{f1} & \textbf{precision} & \textbf{recall} & \textbf{f1} \\
    \hline
    DeBERTa-large       & 0.746 & 0.862 & 0.784 & 0.757 & 0.760 & 0.753 \\
    RoBERTa-MNLI        & 0.650 & 0.772 & 0.691 & 0.834 & 0.842 & 0.834 \\
    DeBERTa-MNLI        & 0.627 & 0.744 & 0.667 & 0.823 & 0.830 & 0.823 \\
    \hline
    
  \end{tabular}
  \caption{Experiments on the second stage of YA-node prediction.}
  \label{tab:ya_relations}
\end{table*}

\begin{table*}[!h]
  \centering
  \begin{tabular}{c|ccc|ccc}
    \hline
    \textbf{Type} & \multicolumn{3}{c|}{\textbf{General Metrics}} & \multicolumn{3}{c}{\textbf{Focused Metrics}} \\
    & \textbf{precision} & \textbf{recall} & \textbf{f1} & \textbf{precision} & \textbf{recall} & \textbf{f1} \\
    \hline
    ARI       & 0.463 & 0.324 & 0.359 & 0.320 & 0.466 & 0.306 \\
    ILO       & 0.542 & 0.499 & 0.514 & 0.564 & 0.646 & 0.594 \\
    \hline
    
  \end{tabular}
  \caption{The result of our submitted system}
  \label{tab:submit}
\end{table*}

\subsection{Experimental Results of S-node Prediction}
First, we artificially generated a certain amount of $r=false$ data in this step and evaluated the impact of this additional data volume. 
Therefore, we performed experiments by controlling the ratio of the amount of $r=false$ data to the amount of $r=true$ data to observe the results. 

Moreover, we experimented with a four-label direct classification model and compared the results with those of the two-step model we ultimately employed.

The results of the first experiment are shown in the appendix~\ref{sec:data_ratio}.
Based on the experimental results, the 1:1 data ratio produced the best outcome. We believe that the 1st-step model only needs to determine whether a relationship exists without considering factors such as the distribution of various relationships that the 2nd-step model should concern. 
Therefore, the 1:1 data ratio makes it easier for the model to distinguish the differences between $r=true$ and $r=false$ data.

The results of the second experiment are shown in Table~\ref{tab:comparison_on_stage1}. Our two-step model framework uses the \textit{DeBERTa-base-1} model, which had the best performance in the first experiment, as the 1st-step model. It can be observed that the models trained directly for four-class classification achieve higher focused scores but have very low general scores. On the other hand, our two-step model achieves a significant improvement in general scores at the expense of sacrificing some focused scores. Overall, the two-step method yields better results.

\subsection{Experimental Results of Y-node Prediction}
We tested the performance of different models in Stage 2. In the experiments of this stage, we trained 12-label classification models. In addition to the training data for the 11 labels extracted from the nodesets, inspired by the experiments in the previous stage, we also included an equal amount of $r=None$ data in training.

The experimental results are shown in Table~\ref{tab:ya_relations}. 
Most of the models had higher Focused scores than General scores.
Among them, DeBERTa-large received the highest General score, whereas RoBERTa-MNLI achieved the highest Focused score.

\subsection{Experimental Results of the Pipelines}
The composition of the pipeline submitted by us in DialAM-2024 is as follows: DeBERTa-base + RoBERTa-MNLI as the first stage model, and DeBERTa-large as the second stage model. The result is shown in Table~\ref{tab:submit}. Our pipeline achieved first place in the ARI Focused score and fourth place in the Global Focused score.

We also modified the models in stage 1 and stage 2 and tested these different pipelines on the test dataset, which was finally released by DialAM-2024. The results are presented in appendix~\ref{sec:ppl}, and we found that we have achieved a much higher score, with the ILO-focused scores surpassing 0.87.

\section{Conclusion}
We propose a two-stage pipeline that predicts argumentative relations and illocutionary relations based on the initial locutions and propositions. This method utilizes data augmentation to optimize the training data and employs a two-step model to predict the relations, incorporating contextual information during prediction. Ultimately, our method achieves good performance in the DialAM24 shared task.

However, due to time constraints and limited computational resources, there are still many aspects of our method that have not been fully optimized. For example, we could appropriately incorporate additional information in locutions to assist the prediction process. It is also worth exploring the possibility of first determining the correspondence between locutions and propositions before predicting the remaining relations. These areas can be further explored and researched.

\section*{Limitations}
In this paper, we design a pipeline that utilizes knowledge of language models, like T5 and DeBERTa, to solve this argument mining problem. For LLMs, we only tested Llama3 (8B)~\cite{llama3modelcard} by fine-tuning a small fraction of parameters. For future works, we can try more LLMs, like Llama2~\cite{touvron2023llama} and Mistral~\cite{jiang2023mistral} with more sizes (e.g., 13B, 70B). Meanwhile, we can augment our argument-mining pipeline with various external knowledge, including commonsense knowledge~\cite{sap2019atomic, do2024constraintchecker, deng2023gold, wang2024candle, wu2023commonsense} event-centric knowledge~\cite{wang2022subeventwriter, wang2023cola, fang2024getting, wang2024abspyramid, wang2024absinstruct, fan2023chain} and factual knowledge~\cite{choi2023kcts}. More importantly, we can also add more modalities like images for relation detection in dialogical argument mining~\cite{zong2023tilfa, shen2024vcd}.

\section*{Acknowledgement}
The authors of this paper were supported by the NSFC Fund (U20B2053) from the NSFC of China, the RIF (R6020-19 and R6021-20), and the GRF (16211520 and 16205322) from RGC of Hong Kong. This paper was also supported by the Tencent AI Lab Rhino-bird Focused Research Program. We also thank the UGC Research Matching Grants (RMGS20EG01-D, RMGS20CR11, RMGS20CR12, RMGS20EG19, RMGS20EG21, RMGS23CR05, RMGS23EG08).

\bibliography{custom}

\appendix

\clearpage

\section{Experiments on different data ratios}
\label{sec:data_ratio}
\begin{table*}
  \centering
  \begin{tabular}{c|ccc|ccc}
    \hline
    \textbf{Model} & \multicolumn{3}{c|}{\textbf{General Metrics}} & \multicolumn{3}{c}{\textbf{Focused Metrics}} \\
    & \textbf{precision} & \textbf{recall} & \textbf{f1} & \textbf{precision} & \textbf{recall} & \textbf{f1} \\
    \hline
    DeBERTa-base-2       & 0.548 & 0.674 & 0.530 & 0.539 & 0.324 & 0.389 \\
    DeBERTa-base-1.5       & 0.550 & 0.672 & 0.536 & 0.506 & 0.290 & 0.358 \\
    DeBERTa-base-1       & 0.541 & 0.671 & 0.507 & 0.526 & 0.332 & 0.397 \\
    \hline
  \end{tabular}
  \caption{Experiments on three different data ratios. }
  \label{tab:data_ratio}
\end{table*}
We conducted experiments using DeBERTa models, with the numbers following the model name indicating the data ratio, i.e., the ratio between the amounts of $r=false$ and $r=true$ data. The results are showin in Table~\ref{tab:data_ratio}.

\section{Full Experiments of Y-node Prediction }
The results are showin in Table~\ref{tab:full_ya_relations}.
Except for the \textit{LLaMa-3-8B} model trained with LoRa, which performed significantly worse, the other models achieved high scores. We speculate that \textit{LLaMa-3-8B} model may not be well-suited for this multi-classification task compared to these smaller models specifically designed for this. 
Most of the models had higher Focused scores than General scores.
Among them, DeBERTa-large received the highest General score, whereas RoBERTa-MNLI achieved the highest Focused score.

\begin{table*}
  \centering
  \begin{tabular}{c|ccc|ccc}
    \hline
    \textbf{Model} & \multicolumn{3}{c|}{\textbf{General Metrics}} & \multicolumn{3}{c}{\textbf{Focused Metrics}} \\
    & \textbf{precision} & \textbf{recall} & \textbf{f1} & \textbf{precision} & \textbf{recall} & \textbf{f1} \\
    \hline
    DeBERTa-large       & 0.746 & 0.862 & 0.784 & 0.757 & 0.760 & 0.753 \\
    LLaMa-3-8B-LoRa     & 0.252 & 0.213 & 0.105 & 0.491 & 0.517 & 0.502 \\
    XLM-RoBERTa-large   & 0.557 & 0.855 & 0.622 & 0.799 & 0.808 & 0.799 \\
    DeBERTa-base        & 0.549 & 0.795 & 0.607 & 0.791 & 0.802 & 0.792 \\
    RoBERTa-MNLI        & 0.650 & 0.772 & 0.691 & 0.834 & 0.842 & 0.834 \\
    DeBERTa-MNLI        & 0.627 & 0.744 & 0.667 & 0.823 & 0.830 & 0.823 \\
    \hline
    
  \end{tabular}
  \caption{Full Experiments on the second stage of YA-node prediction.}
  \label{tab:full_ya_relations}
\end{table*}

\section{Experiments of the pipelines}
\label{sec:ppl}
We modified the model in the second step of stage 1, as well as the model in stage 2, and tested the performance of these different pipelines. The results are shown in the Table~\ref{tab:ppl}.

To our surprise, the second pipeline, DeBERTa-base + RoBERTa-MNLI +  RoBERTa-MNLI, which performed slightly worse on the evaluation set, obtained the highest score in the test set. Its ILO score was significantly higher than the score of the pipeline we submitted. 

We speculate that this might be because our evaluation set consisted of only 78 randomly selected nodesets from the training dataset, which could have significant differences in data distribution and relationship distribution compared to the final test set. As a result, the pipeline that performed best on the validation set may have had poorer performance on the test set, while some pipelines that performed slightly worse on the validation set happened to achieve better scores on the test set.
\begin{table*}
  \centering
  \begin{tabular}{c|c|ccc|ccc}
    \hline
    \textbf{Model} & \textbf{Type} &\multicolumn{3}{c|}{\textbf{General Metrics}} & \multicolumn{3}{c}{\textbf{Focused Metrics}} \\
    & & \textbf{precision} & \textbf{recall} & \textbf{f1} & \textbf{precision} & \textbf{recall} & \textbf{f1} \\
    \hline
    DeBERTa-base + RoBERTa-MNLI & ARI  &  \textbf{0.463} & 0.324 & \textbf{0.359} & 0.320 & \textbf{0.466} & \textbf{0.306} \\
    + DeBERTa-large (\textbf{submitted})   &  ILO & 0.542 & 0.499 & 0.514 & 0.564 & 0.646 & 0.594 \\

    \hline
    DeBERTa-base + RoBERTa-MNLI & ARI & 0.463 & 0.324 & 0.359 & 0.320 & 0.466 & 0.306 \\
    + RoBERTa-MNLI & ILO & 0.660 & \textbf{0.796} & \textbf{0.705} & \textbf{0.873} & \textbf{0.902} & \textbf{0.883} \\
    \hline
    DeBERTa-base + DeBERTa-large & ARI & 0.366 & \textbf{0.469} & 0.331 & \textbf{0.393} & 0.261 & 0.285 \\
    + DeBERTa-large  & ILO & \textbf{0.676} & 0.763 & 0.703 & 0.662 & 0.648 & 0.652 \\
    \hline
    
  \end{tabular}
  \caption{Experiments of different pipelines.}
  \label{tab:ppl}
\end{table*}

\section{Additional Data Description}
\label{sec:data_desc}
Our dataset comprises a total of 1,478 nodesets. We randomly selected 78 nodesets as the evaluation set, leaving the remaining 1,400 nodesets for the training set. 

The training set contains 5,365 RA data samples, 1,181 CA data samples, 5,596 MA data samples, and 32,626 YA data samples. In the evaluation set, there are 268 RA data samples, 59 CA data samples, 279 MA data samples, and 1,631 YA data samples.

The selected 78 nodesets are:
'nodeset18321', 'nodeset21402', 'nodeset21463', 'nodeset23939', 'nodeset18455', 'nodeset19912', 'nodeset23828', 'nodeset21575', 'nodeset17918', 'nodeset23771', 'nodeset21041', 'nodeset18846', 'nodeset18850', 'nodeset23887', 'nodeset18775', 'nodeset21044', 'nodeset18877', 'nodeset23794', 'nodeset23512', 'nodeset25524', 'nodeset21390', 'nodeset23605', 'nodeset23769', 'nodeset23526', 'nodeset17938', 'nodeset19911', 'nodeset20342', 'nodeset21438', 'nodeset18311', 'nodeset19159', 'nodeset19742', 'nodeset23547', 'nodeset18764', 'nodeset21384', 'nodeset21294', 'nodeset19153', 'nodeset20755', 'nodeset23869', 'nodeset17923', 'nodeset20303', 'nodeset23894', 'nodeset23715', 'nodeset23484', 'nodeset20332', 'nodeset23505', 'nodeset21577', 'nodeset21595', 'nodeset19341', 'nodeset21023', 'nodeset23746', 'nodeset20871', 'nodeset25400', 'nodeset18271', 'nodeset20343', 'nodeset21473', 'nodeset21571', 'nodeset25691', 'nodeset21452', 'nodeset18848', 'nodeset23721', 'nodeset18794', 'nodeset25522', 'nodeset25499', 'nodeset21393', 'nodeset17940', 'nodeset23876', 'nodeset23927', 'nodeset23498', 'nodeset23900', 'nodeset19095', 'nodeset20981', 'nodeset21603', 'nodeset21451', 'nodeset18266', 'nodeset25754', 'nodeset19091', 'nodeset23859', 'nodeset23834'

\end{document}